\documentclass[11pt,a4paper]{article}
\usepackage{times,latexsym}
\usepackage{url}
\usepackage[T1]{fontenc}

   \usepackage[acceptedWithA]{tacl2018v2}
\usepackage[]{tacl2018v2}

\usepackage{microtype}
\usepackage{booktabs}
\usepackage{comment}
\usepackage{pifont}
\usepackage{xspace}
\usepackage{ulem}
\usepackage{amssymb}
\usepackage{tabularx}
\usepackage{multirow}
\usepackage{subcaption}
\usepackage{graphicx}

\usepackage{placeins}  %

\DeclareUnicodeCharacter{00B3}{\textsuperscript{3}}

\newcommand{\corpus}{\textsc{MultiCite}}

\newcommand{\cmark}{\ding{51}}%
\newcommand{\xmark}{\ding{55}}%

\title{Doing it right: It's time for addressing issues in citation context analysis}
\title{\corpus{}: \\It's Time for Addressing Issues in Citation Context Analysis}
\title{\corpus{}: \\ Revisiting Citation Context Analysis in Natural Language Processing}
\title{\corpus{}: Modeling realistic citations requires \\ moving beyond the single-sentence single-label setting}

\author{
 Anne Lauscher\thanks{\xspace \xspace Part of the work was conducted during an internship at the Allen Institute for AI.} \textsuperscript{1} \hspace{1.4em} Brandon Ko\textsuperscript{2} \hspace{1.4em}  Bailey Kuehl\textsuperscript{3} \hspace{1.4em} \textbf{Sophie Johnson}\textsuperscript{3}  \\ \textbf{David Jurgens\textsuperscript{4}} \hspace{1.4em} \textbf{Arman Cohan\textsuperscript{3}} \hspace{1.4em}  \textbf{Kyle Lo\textsuperscript{3}} \vspace{4pt}\\ 
 \textsuperscript{1}Data and Web Science Research Group, University of Mannheim, Germany \\
 \textsuperscript{2}University of Washington, Seattle WA \\
 \textsuperscript{3}Allen Institute for AI, Seattle WA \\
 
 \textsuperscript{4}School of Information, University of Michigan, Ann Arbor MI \\
\footnotesize{\texttt{
    anne@informatik.uni-mannheim.de \hspace{1.4em} bk36@cs.washington.edu}}\vspace{-4pt}\\ \footnotesize{\texttt{\{baileyk,sophiej,armanc,kylel\}@allenai.org \hspace{1.4em} jurgens@umich.edu}} \\
}

\date{}

\begin{document}
\normalem
\maketitle
\begin{abstract}
Citation context analysis (CCA) is an important task in natural language processing that 
studies
\emph{how} and \emph{why} scholars discuss each others' work. 
Despite decades of study, traditional frameworks for CCA have largely relied on overly-simplistic assumptions of how authors cite, which ignore several important phenomena.
For instance, scholarly papers often contain rich discussions of cited work that span multiple sentences and express multiple intents concurrently.
Yet, CCA is typically approached as a single-sentence, single-label classification task, and thus existing datasets fail to capture this interesting discourse. 
In our work, we 
address this research gap by proposing a novel framework for CCA as a \emph{document-level context extraction and labeling} task.
We release \corpus{}, a new dataset of 12,653 citation contexts from over 1,200 computational linguistics papers.
Not only is it the largest collection of expert-annotated citation contexts to-date, \corpus{} contains multi-sentence, multi-label citation contexts within full paper texts.
Finally, we demonstrate how our dataset, while still usable for training classic CCA models, also supports the development of new types of models for CCA beyond fixed-width text classification.
We release our code and dataset at \href{https://github.com/allenai/multicite}{https://github.com/allenai/multicite}.
\end{abstract}

\section{Introduction}
In scientific writing, citations are answers to questions. Citing authors preemptively respond to questions such as---\emph{why is this work needed}, \emph{where has a fact been previously shown}, or \emph{what technique is used}---in order to construct and justify an argument about the correctness of the claims in their work~\citep{10.2307/284636, teufel2014scientific}. Together, these citations connect the current paper to the broader discourse of science~\citep[e.g.,][]{garfield1955citation,siddharthan-teufel-2007-whose}, help signal future impact and uses~\citep[e.g.,][]{10.1002/asi.23612}, and, in downstream applications, can aid in summarizing a work's contributions~\citep[e.g.,][]{qazvinian-radev-2008-scientific, lauscher2017university}. However, the function a particular citation serves---what questions it answers---is often implicit and relies on the reader to understand that citation's purpose. Here, we introduce a new dataset and formalism of citation function, showing that a citation can answer multiple questions simultaneously, that the citation context answering these questions often extends beyond a single sentence, and that recognizing a citation's intent can be reframed as a question-answering task, unifying its place within larger reading comprehension tasks.

The importance and role of citations in understanding scholarly work has been recognized across multiple disciplines, from sociology~\citep[e.g.,][]{garfield1964use, garfield1970citation} to computer science~\citep[e.g.,][]{10.1002/asi.23612, yasunaga2019scisummnet}. Prior computational work has attempted analyze citations through classifying aspects of what questions they answer, e.g., their function~\citep[e.g.,][]{teufel_automatic_2006,jurgens2018measuring}, sentiment~\citep[e.g.,][]{athar_sentiment_2011,jha_nlp-driven_2016}, or centrality to a paper~\citep{valenzuela2015identifying}. However, these works have varied significantly in how they define a citation's contexts---i.e., the text relating to a citation necessary to understand its intent---with most only examining a single sentence~\citep[e.g.,][]{athar_sentiment_2011, dong_ensemble_2011,ravi2018article, cohan2019structural} and some leaving the context notion undefined~\citep{jurgens2018measuring, vyas2020article}. Further, few have recognized that often a citation's purpose is not singular, that the citation may be simultaneously answering multiple questions. 
We revisit these assumptions to show the complexity of citations' discourse functions.

\paragraph{Contributions.} Our contributions are three-fold: \textbf{(i)} We first demonstrate the existence of the described phenomena and discuss their importance in a qualitative analysis of NLP publications. Building upon the identified issues, we then propose a new framework for citation context analysis (CCA). \textbf{(ii)} In order to allow for training models under our proposed framework, we next present our new \textbf{Multi}-Sentence \textbf{Multi}-Intent \textbf{Cit}ation~(\corpus{}) corpus, a carefully-annotated collection of full-text publications with citation context and citation intent information. \corpus{} substantially advances our ability to perform CCA by being the only dataset which captures multi-sentence and multi-intent citation contexts. \textbf{(iii)} Finally, we establish a range of computational baselines on our new data set under %
two different task formulations: We start with intent classification as the most traditional setup and demonstrate the importance of using gold citation contexts. %
Then, we reframe understanding citation function as a form of question answering: Given a pair of papers, what is the reasoning behind the citation?  %
We hope that our work fuels and inspires more research on accurate CCA models, and enables computational linguists to study more complex and overlooked phenomena of citation usage.

\section{Related Work}
\setlength{\tabcolsep}{4pt}
\begin{table*}[!htp]\centering
\small
\begin{tabular}{llllc}\toprule
\textbf{Author/ Year} &\textbf{Concept} &\textbf{Size} &\textbf{Context notion?} &\textbf{Multi-label?} \\\midrule
\citet{pride2020authoritative} &Purpose \& Influence &11,233 &Single sentence & \xmark\\
\citet{vyas2020article} & Sentiment & Reanno. &No context annotated &\xmark\\
\citet{tuarob2019automatic} &Function (Algorithm) &8,796 &3 Sentences &\xmark \\
\citet{cohan2019structural} &Intent &9,159+1,861 &Single sentence &\xmark \\
\citet{ravi2018article} &Sentiment &8,925 &Single sentence &\xmark \\
\citet{jurgens2018measuring} &Function &1,969 &No context annotated &\xmark \\
\citet{jha_nlp-driven_2016} &Purpose \& Polarity &3,271 &Flexible within 4 sentences &\xmark \\
\citet{valenzuela2015identifying} &Meaningfulness &465 &? &\xmark \\
\citet{abu-jbara_purpose_2013} &Purpose \& Polarity &3,271 &Flexible within 4 sentences &\xmark \\
\citet{li-etal-2013-towards} &Function &? &? &? \\
\citet{jochim_towards_2012} &Citation Facets &2,008 &No context annotated &\xmark \\
\citet{athar-teufel-2012-context} &Sentiment &1,741 &Flexible within 4 sentences & 1 label/sent. \\
\citet{dong_ensemble_2011} &Function \& Polarity &? & Single sentence &\xmark \\
\citet{athar_sentiment_2011} &Sentiment &8,736 &Single sentence &\xmark\\
\citet{teufel_automatic_2006} &Function &548 &? &\xmark \\
\midrule
\textsc{MultiCite} (this work) & Function & 12,653 & Flexible & \cmark\\
\bottomrule
\end{tabular}
\caption{Existing citation context analysis data sets with their properties in comparison to this work.}\label{tab:rw}
\end{table*}

We describe traditional works on CCA with respect to task forms and published resources. An overview of these works is provided in Table~\ref{tab:rw}.\footnote{For detailed reviews of CCA, we refer to \citep{iqbal2020decade} and \citep{hernandez-alvarez_gomez_2016}.}

The problem of citation analysis dates back to the seminal work of \citet{teufel_automatic_2006}, who addressed the task as a multi-class classification problem without a clear notion of the citation context. Building upon this, researchers initially operated on a single-sentence citation context~\citep[e.g.,][]{athar_sentiment_2011, dong_ensemble_2011}, but later acknowledged the importance of precise and flexible context windows~\citep[e.g.,][]{athar-teufel-2012-context, abu-jbara_purpose_2013}. Most recently, in-line with the advent of deep neural model architectures, research efforts focused on increasing the size of the published resources~\citep{cohan2019structural, tuarob2019automatic, pride2020authoritative}. However, larger data sets came at the expense of annotating precise context notions, which naturally leads to increased complexity of the annotation task.
In \S\ref{sec:framework}, we demonstrate the importance of flexible contexts and further show that contexts expressing multiple intents exist. None of the preceding works provides a multi-label data set: though \citet{athar-teufel-2012-context} allow for multiple labels around a citation marker, they assign exactly one label to each sentence within the context. In contrast, we present a new framework for CCA together with the largest, flexible-context, and only multi-label annotated resource for CCA to-date.

\section{Multi-Sentence Multi-Intent Framework for Citation Analysis}
\label{sec:framework}
We recap the framework under which preceding works conducting citation context analysis (CCA) have been operating. Based on this, we analyze its shortcomings and propose to move to our multi-sentence multi-intent framework.

\paragraph{Traditional Citation Analysis Framework.} As outlined before, preceding works have mostly been focusing on single-sentence single-intent citation context classification~\citep[e.g.,][\emph{inter alia}]{athar_sentiment_2011,jurgens2018measuring, cohan2019structural,pride2020authoritative}. Accordingly, these works can be subsumed under a single notion, which we refer to as the \emph{traditional framework} of CCA:

Let $B$ be a target paper cited by a citing paper $A=[s_0,...,s_N]$, which corresponds to a sequence of $N$ sentences $s_i$, and let $c$ be the corresponding citation marker (a single token, e.g., \emph{[1]}), indicating the citation of $B$ in $A$. The citation context corresponds to a single sentence $s_c$ with $c \in s_c$, often referred to as \emph{the citance}~\citep{nakov2004citances}. Then, given a set of labels $L=\{l^{(j)}\}^{M}_{j = 1}$ of size $M$, the output is a single label $l^{(j)}$ assigned to $s_c$.

\paragraph{Overseen Phenomena.} We claim that the traditional framework does not account for two important phenomena: multi-sentence contexts, and multi-label intents. We will support our claim by providing examples, which we discuss under a citation labeling scheme inspired by \citet{jurgens2018measuring}, depicted in Table~\ref{tab:scheme}.\footnote{We later adopt this labeling scheme for the purpose of creating our data set.} 

\vspace{0.75em}
\noindent\emph{Multi-Sentence.} First, consider the following citance $s_c$ with the underlined citation marker $c$: 

\vspace{0.75em}
\parbox{0.9\linewidth}{\emph{``\uline{(Gliozzo et al., 2005)} succeeded eliminating this requirement by using the category name alone as the initial keyword, yet obtaining superior performance within the keywordbased approach.''}}

\vspace{0.75em} 
\noindent This citance alone provides background information about a previous approach. Consequently, one would label it as \emph{Background}. However, in the subsequent sentence, we can read the following:

\vspace{0.75em}
\parbox{0.9\linewidth}{\emph{``The goal of our research is to further improve the scheme of text categorization from category name, which was hardly explored in prior work.''}}

\vspace{0.75em} 
\noindent Only in this sentence, the underlying intent of the authors of $A$ is exposed: the cited publication $B$ is used as a \emph{Motivation} for the presented research. Under the traditional framework, we fail to correctly classify this example.\footnote{We acknowledge that positional features could help. In this work, however, we focus on textual semantics only.}

\vspace{0.75em}
\noindent\emph{Multi-Intent.} Next, consider the following citance $s_c$ with the underlined citation marker $c$: 

\vspace{0.75em}
\parbox{0.9\linewidth}{\emph{``In our experiments we use the same definition of structural locality as was proposed for the ISBN dependency parser in \uline{(Titov and Henderson, 2007b)}.''}}

\vspace{0.75em}
\noindent Under the citation labeling scheme inspired by \citet{jurgens2018measuring}, depicted in Table~\ref{tab:scheme}, %
this sentence can be labeled as \emph{Similarities}. However, another possibility is to label this citance as \emph{Uses}, as the authors are adopting a definition of the cited work. In some of the preceding works, ambiguous citances were reportedly removed~\citep[e.g.,][]{cohan2019structural}, leading to an artificial simplification of the task.

\vspace{0.75em}
\noindent\emph{Multi-Sentence Multi-Intent.} While we noticed that the two phenomena outlined above exist in isolation, we also observe instances which combine both: 

\vspace{0.75em}
\parbox{0.9\linewidth}{\emph{``Results Table 1 compares the published BERT BASE results from \uline{Devlin et al. (2019)} to our reimplementation with either static or dynamic masking. We find that our reimplementation with static masking performs similar to the original BERT model, and dynamic masking is comparable or slightly better than static masking.''}}

\vspace{0.75em}
\noindent Here, the published results from the well-known BERT paper, a research artefact, are \emph{Used} as a baseline ($s_c$). Then, the authors compare their reimplementation as well as their extension to these scores (sentence 2), resulting in expressed \emph{Similarities} as well as \emph{Differences}.

\paragraph{Multi-Sentence Multi-Intent CCA.} As discussed, all the examples above exhibit phenomena, which can not be fully expressed under the traditional framework. While we acknowledge that instances falling under this category are less common, we argue that the community should not longer ignore this ``long tail'' and address the phenomena highlighted above. To this end, we propose a new multi-sentence multi-intent framework for CCA:

Let $B$ be a target paper cited by a citing span of text $T=[s_0,...,s_N]$, which corresponds to a sequence of $N$ sentences $s_i$ in a citing paper $A$. Given a set of labels $L=\{l^{(j)}\}^{M}_{j = 1}$ of size $M$, the output consists of \emph{all} (dis)continuous sequences $E=[s_{e0},...,s_{eO}]$ of size $O$, with $1 <= O <= N$ (the set of elements on $E$ build a subset of the set of elements in $T$), which contain \emph{all} evidentiary sentences $s_e$ for at most $k<=M$ citation intents $l^{(j)}$ expressed in this context. %

Under this notion, we can still instantiate the traditional framework and approach CCA as a single-sentence multi-class classification task by restricting $T$ to only contain $s_c$, i.e., the citance, for any citation marker $c$ in $A$ and restricting $k$ to $k=1$, i.e., assigning a single label $l^{(j)}$ only. However, aiming towards a more holistic picture, we are longing to set $T=A$, thereby feeding the whole full text of a paper $A$. Similarly, we want to set $k=M$, thereby allowing for full multi-intent classification. By setting $T=A$ and $k=M$ we cast the task as full-blown multi-sentence multi-intent CCA, returning all underlying intents of $B$ being cited by $A$ with all evidentiary contexts.

\section{\corpus{}: A New Corpus for Citation Analysis}
    \begin{table*}[t]
    \centering
    \small{
    \begin{tabularx}{\textwidth}{lX}%
    \toprule
         \textbf{Intent} & \textbf{Description} \\
         \midrule
         \emph{Background} & The target paper provides relevant information for this domain.  \\
         \emph{Motivation} & The target paper provides motivation for the source paper. For instance, it illustrates the need for data, goals, methods etc.   \\
         \emph{Uses} & The source paper uses an idea, method, tool, etc. of the target paper.  \\
         \emph{Extends} & The source paper extends an idea, method, tool, etc. of the target paper.  \\
         \emph{Similarities} & The source paper expresses similarities towards the target paper. Either similarities between the source and the target paper or similarities between another publication and the target paper.  \\
         \emph{Differences} & The source paper expresses differences towards the target paper. Either differences between the source and the target paper or differences between another publication and the target paper.  \\
         \emph{Future Work} & The target paper is a potential avenue for future research.  Often corresponds to hedging or speculative language about work not yet performed.  \\
         \bottomrule
    \end{tabularx}}
    \caption{Our citation intent labeling scheme based on \citet{jurgens2018measuring}.}
    \label{tab:scheme}
\end{table*}

As it can be seen from the discussion of the related work, to-date there exists no single data set which allows for training models under our proposed framework. We close this research gap and present \corpus{}, the first multi-sentence multi-intent corpus for citation analysis consisting of $1,200$ computational linguistics publications.
\subsection{Annotation Study}
We describe the creation of \corpus{}.

\paragraph{Sampling Procedure.} 

We procure an initial corpus of candidate papers from S2ORC~\cite{lo-etal-2020-s2orc}, a large collection of open-access full-text papers annotated with inline citation mentions resolved to their cited papers, and filtering to 50K papers from the ACL Anthology or from arXiv with a \textsc{cs.CL} category.  %

To capture interesting cases, which, ideally, exhibit the phenomena we are targeting with our research, we initially experimented with several paper sampling strategies and found the following strategy to yield a variety of  interesting publication pairs: we compute the number of paragraphs in which the target marker appears divided by the total number of paragraphs and retrieve the top $k$ papers. This way, we capture publication pairs, where the target paper is cited many times and therefore, in many different ways. We hypothesize that these papers  play a central role in the citing paper.

\paragraph{Mention Annotation.} To guide our annotators in finding passages in the citing paper A which are talking about the cited publication B, we highlight mentions of B in A. We automate this by highlighting all markers given in the respective S2ORC file. To compute the reliability of this method, we let annotators manually identify all references to B including citation markers, scientific entity names, and other co-references such as \emph{``The authors ...''} in a small sample of 262 publication pairs. We then compute the agreement with the gamma tool~\cite{mathet-etal-2015-unified} and obtain a mean score of 0.60 gamma macro averaged over the publications. We therefore explicitly instruct our annotators to use the highlighting as a rough guidance but to manually check for other mentions and co-references.

\begin{figure*}
    \centering
    \includegraphics[width=\linewidth]{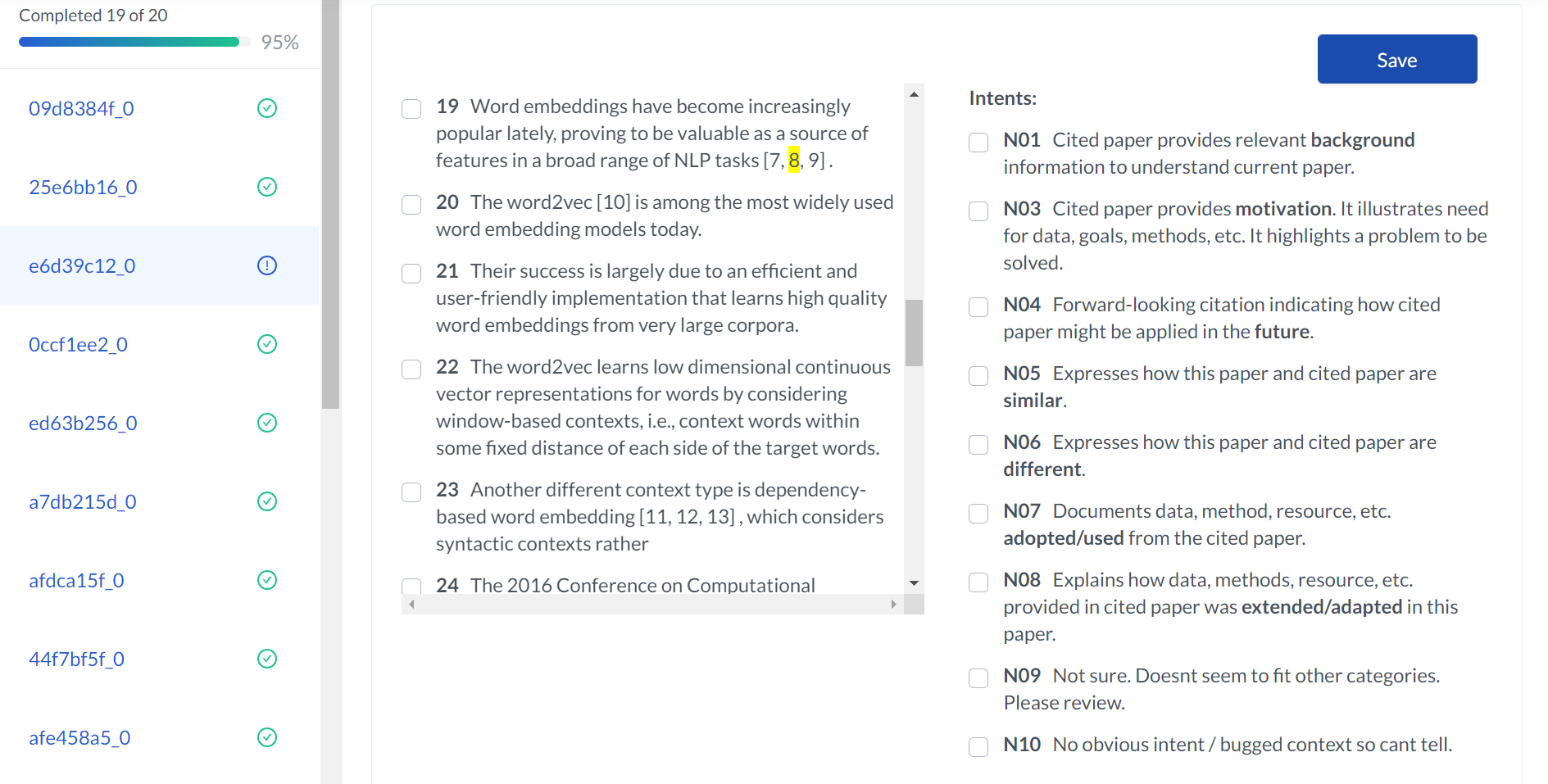}
    \caption{The interface of our dedicated annotation platform: on the left hand side, the annotator can browse through their assigned publications; in the center, each sentence (choosable via checkboxes) of the citing publication is displayes (mentions of the cited publication are highlighted in yellow); on the right hand side, applying intents can be selected.}
    \label{fig:screenshot}
\end{figure*}

\paragraph{Annotation Task, Scheme, and Platform.} 
 Our annotation task consists of two main steps: 
(1) Given a paper $A$ and a paper $B$ (identified via some marker), identify all citing contexts. (2) Given each individual citation context, assign a label to the context answering the question why $A$ is mentioning $B$. 
Aiming to improve upon the issues outlined before, we make sure that our guidelines include \emph{explicit} as well as \emph{implicit} citations (proxied via a coreference, e.g., via pronoun or name) into our task. Regarding the label scheme, we focus on intent classification and start from the annotation scheme of \citet{jurgens2018measuring}, which we chose due to its relative simplicity compared to the one of \citet{teufel_automatic_2006}. We then iteratively adapted the guidelines according to the discussions with the annotators. In the end, we only split the \emph{Comparison or Contrast} class from the original scheme into two classes, \emph{Similarities} and \emph{Differences}. The complete scheme is depicted in Table~\ref{tab:scheme}. 

To facilitate the process as far as possible, we developed a dedicated annotation platform. A screenshot of the interface is provided in Figure~\ref{fig:screenshot}. 

\paragraph{Annotation Process.} For our annotation study, we hired nine graduate students in NLP recruited via Upwork.  With each of them, we conduct an hour of one-on-one training. The annotators independently completed an hour of annotations, which were manually reviewed and used for a second hour of one-on-one training focused on feedback and correcting common mistakes.  Annotators were then allowed to work independently on batches of 20 papers at a time with manual annotation review after each batch for quality control.
Annotators were encouraged to indicate ``Unsure'' for citation contexts with ambiguous labels, and leave comments describing their thoughts. For these cases, two of the nine students were recruited to do a subsequent adjudication round resolving to one or more existing labels, if possible, else leaving as ``Unsure''  if unable to come to consensus.  Annotators were paid between \$25-35 USD per hour, based on their indicated offer on Upwork. 

\paragraph{Inter-Annotator Agreement} 

Producing a single measure of inter-annotator agreement (IAA) is difficult for data collected in this manner.  Annotators might agree on the same intents but disagree on which sentences belong in the citation context, vice versa, or disagree on both fronts. While some prior work has developed IAA measures that capture both context selection and labeling, e.g., $\gamma$ by \citet{mathet-etal-2015-unified}, such methods aren't widely adopted in NLP and thus resulting IAA values can be difficult to interpret.  We opt instead to report two measures: (a) mean accuracy of humans predicting the intent labels given gold contexts, and (b) mean F1 score of humans predicting the context sentences given gold intents.

For (a), we sample a set of 54 gold intent-context pairs across 5 papers.  For each example, two annotators who haven't seen these papers previously were shown the gold context and asked to select all possible intents from 8 categories (including an ``Unsure'' option).  Mean accuracy is 0.76 when counting \emph{any} gold label match as correct, and 0.70 when only counting cases when \emph{all} predicted labels match the gold annotations as correct.  

For (b), we sample 120 single gold intent-context pairs, each from a different annotated paper.  For each example, two annotators who haven't seen these papers previously were shown the gold intent and asked to select the context sentences from among a randomly chosen window of 20 sentences encapsulating the gold context.  Mean sentence-level F1 scores are 0.64, 0.63 and 0.65, respectively for gold contexts of length 1, 2 or 3+ sentences.

\subsection{Corpus Analysis}
\begin{figure*}[t]
     \centering
     \begin{subfigure}[t]{0.24\textwidth}
         \centering
         \includegraphics[width=1.0\linewidth,trim=0.0cm 0cm 0cm 0cm]{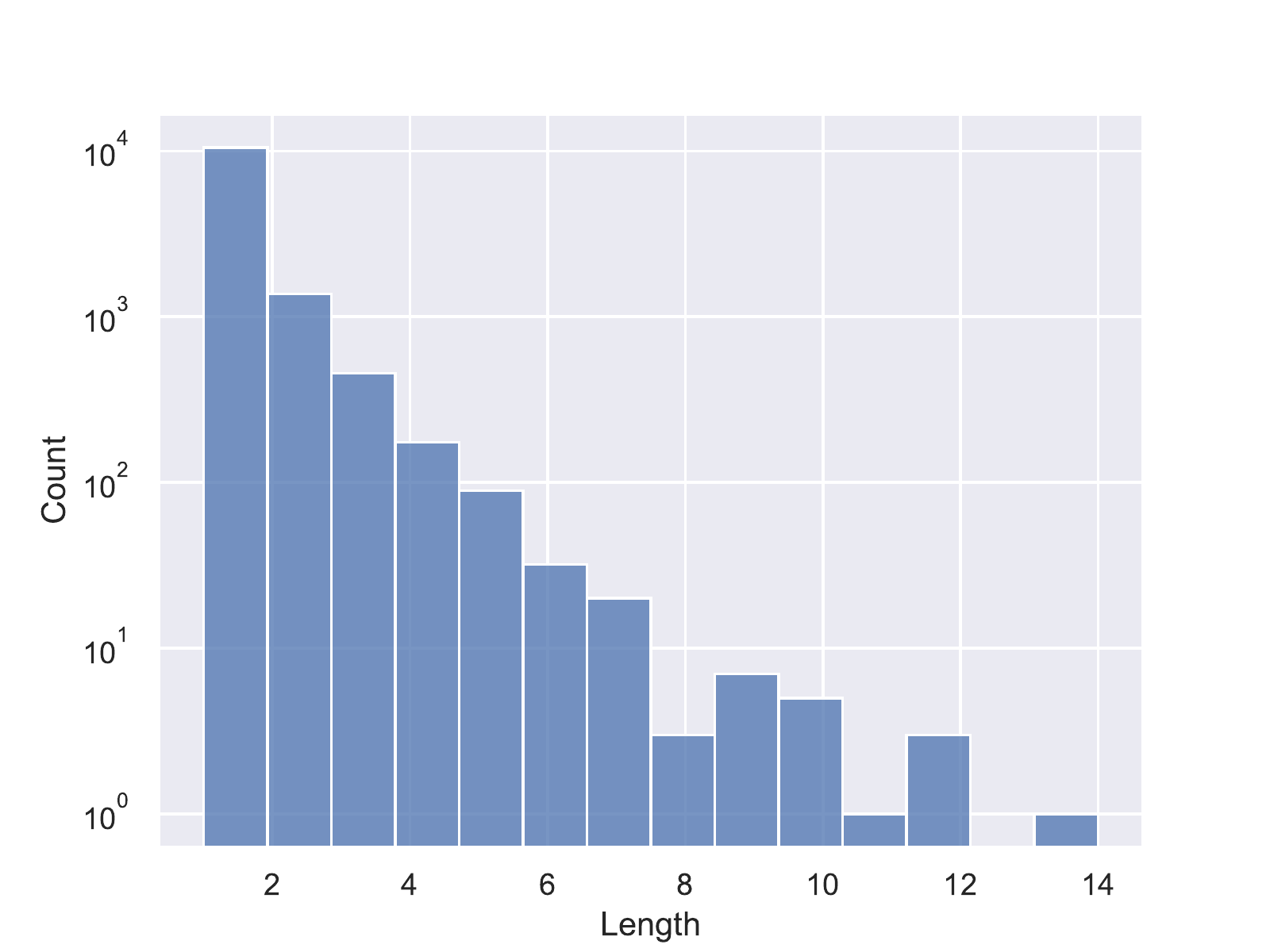}
         \caption{\footnotesize Context length distribution.}
         \label{fig:context_length}
    \end{subfigure}
    \begin{subfigure}[t]{0.24\textwidth}
         \centering
         \includegraphics[width=1.0\linewidth,trim=0.0cm 0cm 0cm 0cm]{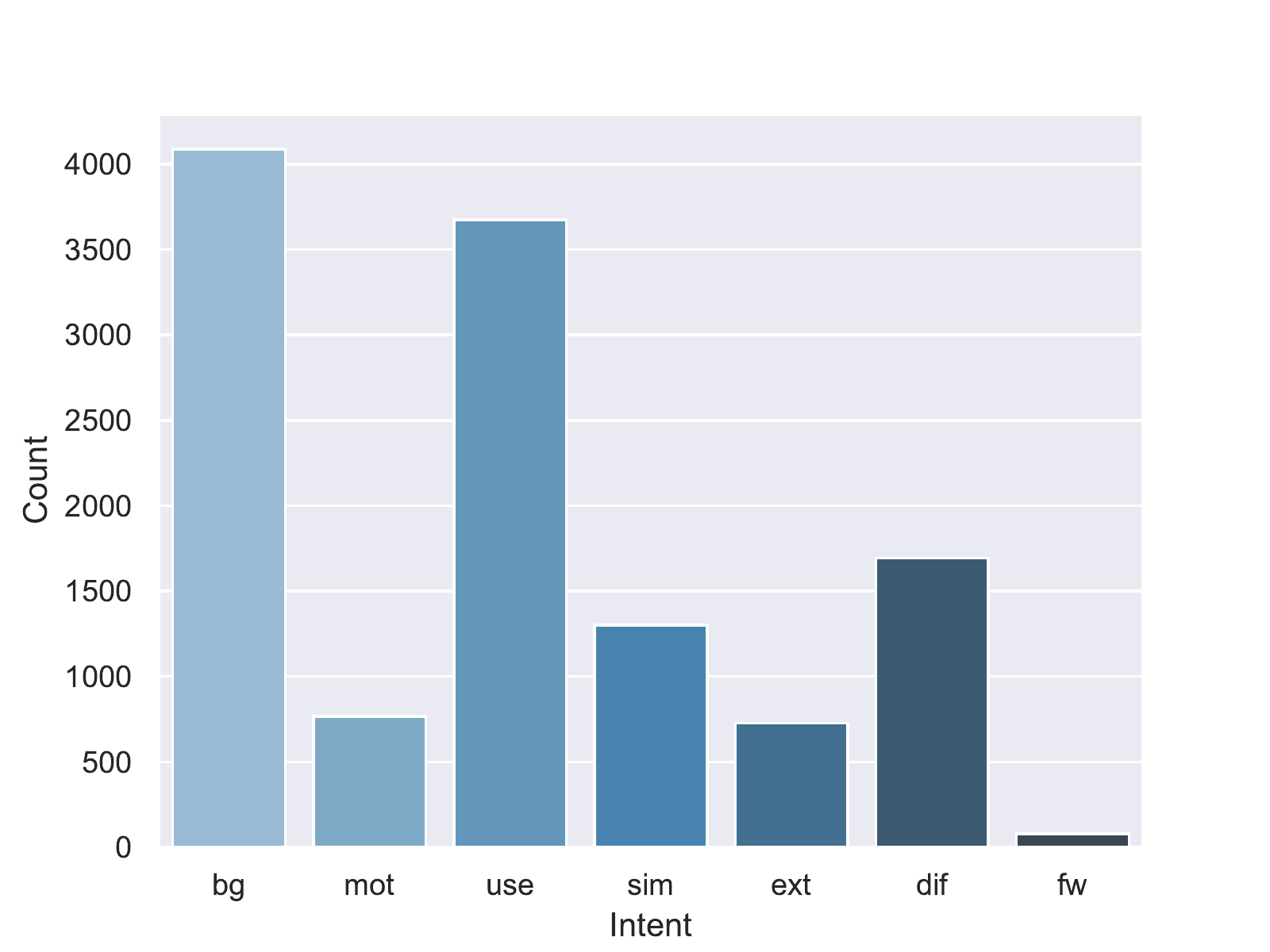}
         \caption{\footnotesize Intent distribution.}
         \label{fig:intents}
     \end{subfigure}
    \begin{subfigure}[t]{0.24\textwidth}
         \centering
         \includegraphics[width=1.01\linewidth,trim=0.0cm 0cm 0cm 0cm]{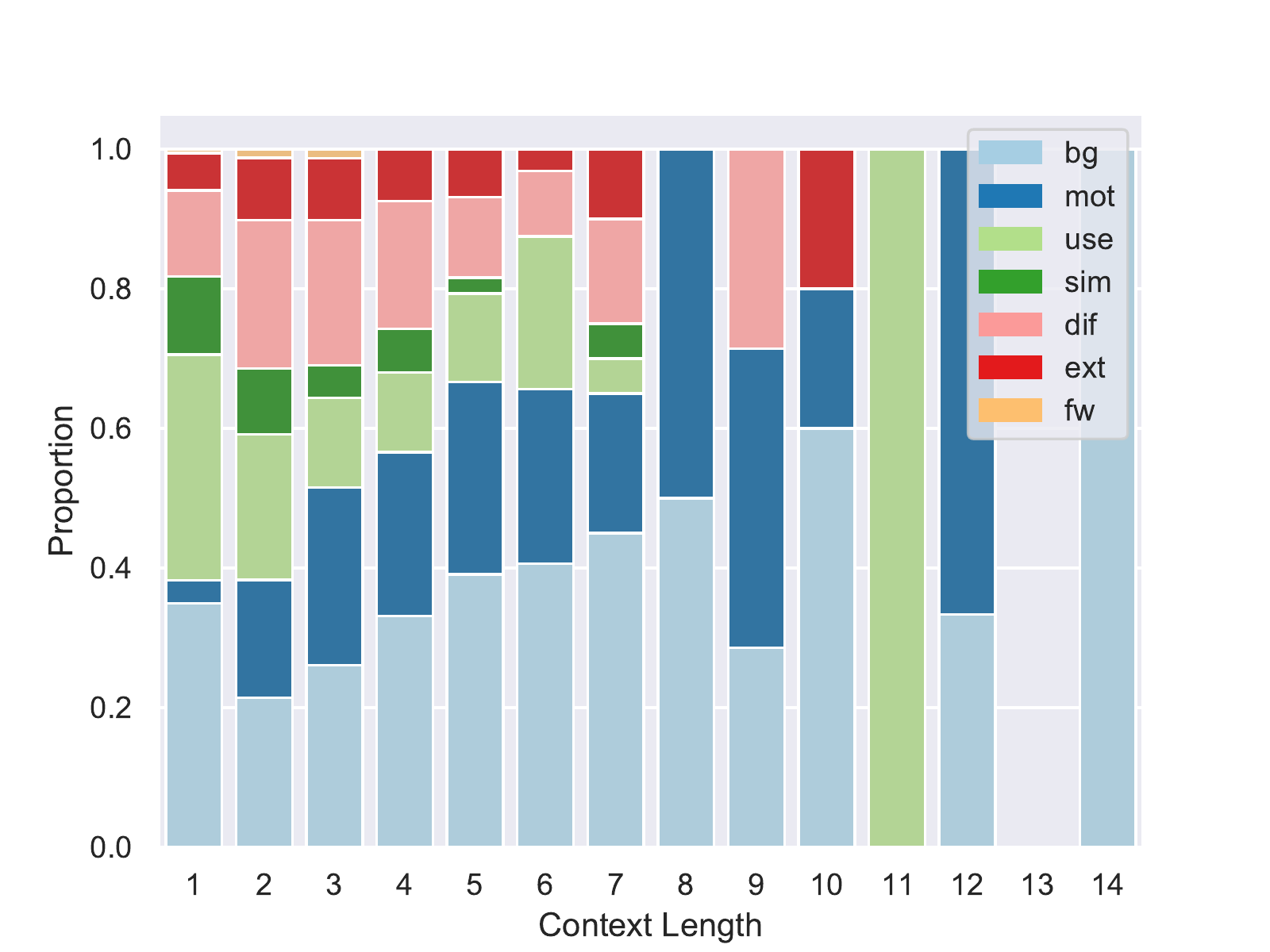}
         \caption{\footnotesize Context length per intent.}
         \label{fig:length_intent}
     \end{subfigure}
    \begin{subfigure}[t]{0.24\textwidth}
         \centering
         \includegraphics[width=1.0\linewidth,trim=0.0cm 0cm 1.5cm 0cm]{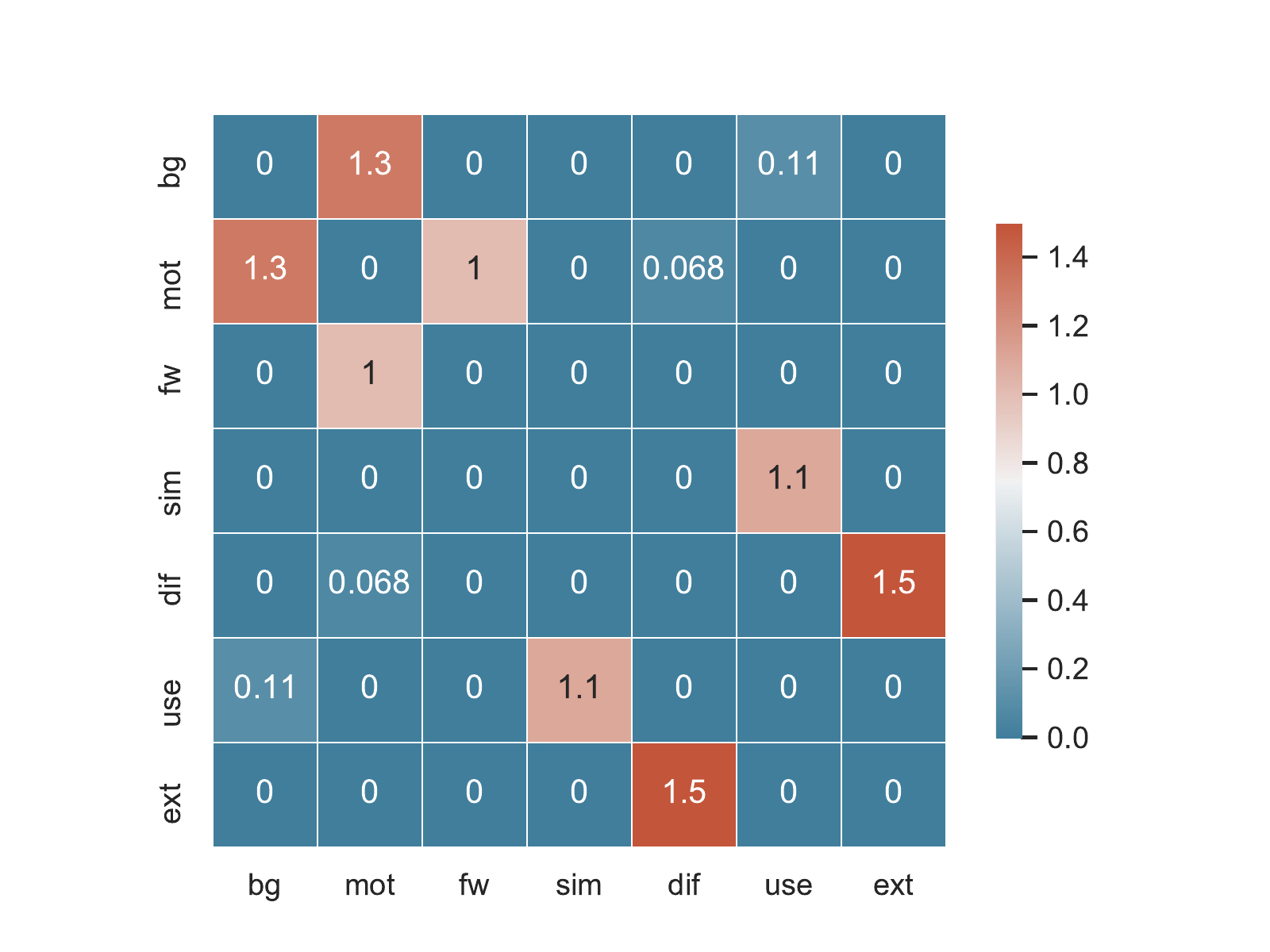}
         \caption{\footnotesize PMI between intents.}
         \label{fig:pmi}
     \end{subfigure}
        \caption{Results of the quantitative corpus analysis: (a) distribution of context lengths (log scale), (b) distribution of intents, (c) distribution of context lengths per intent class, (d) pointwise mutual information (PMI) between intent classes. We use the following intent abbreviations: \emph{Background} (bg), \emph{Motivation} (mot), \emph{Uses} (use), \emph{Similarities} (sim), \emph{Extends} (ext), \emph{Differences} (diff),  \emph{Future Work} (fw).}
        \label{fig:descriptive}
        \vspace{-0.5em}
\end{figure*}
Our new \corpus{} consists of 1,193 publications, which are annotated with in total 12,653 citation contexts representing 4,555 intents per publication. To obtain a deeper insight into the nature of our annotations, we show the results of a more extensive quantative analysis in Figure~\ref{fig:descriptive}.

Unsurprisingly, the context length distribution (Figure~\ref{fig:context_length}) indicates that most annotations consist of single-sentence contexts only. However, a substantial number of contexts went beyond the single sentence. Indeed, we were able to capture contexts consisting of up to 14 sentences, which highlights that artificially restricting the context length will lead to loss of potentially relevant information. 

The intent distribution (Figure~\ref{fig:intents}) is highly skewed. This is in-line with results reported in previous work~\citep{jurgens2018measuring}: the most dominant class is \emph{Background (bg)}, followed by \emph{Uses (use)}, while \emph{Future work (fw)} occurs least often. 

The distribution of context lengths per intent class~(Figure~\ref{fig:length_intent}) reveals that each of the intents can be expressed with a one to three sentences only. However, for instance for \emph{Motivation (mot)}, contexts clearly exhibit a variety in length. We hypothesize that this intent is of higher complexity and therefore often requires more context. 

Finally, we compare the label co-occurrences. To this end, we compute pointwise mutual information (PMI) between the intents (Figure~\ref{fig:pmi}). The highest PMI is observed between \emph{Extends (ext)} and \emph{Differences (dif)}, which points to the strong interrelation between those two classes: when authors introduce an extension they made to a research artefact from preceding work, they often express at the same time, that exactly this aspect makes the work different from what was previously presented.

\section{Importance of Multi-Sentence Contexts}
In order to assess the importance of the correct multi-sentence citation context for an intent classification, we conduct a series of experiments aligned with previous works~\citep[e.g.,][]{jha_nlp-driven_2016}, in which we feed various amounts of context sentences to a multi-label classification model.

\subsection{Experimental Setup}
We describe the experimental setup for our multi-label classification experiments.

\paragraph{Standard Split.} For all our experiment, we use the same standard split on the publication-level of \corpus{}, to not leak any information from the articles. Concretely, $30\%$ of the publications are reserved for testing, $70\%$ for training, from which, in turn, $30\%$ are reserved for model validation. This results in 5,491 training instances, 2,447 development instances, and 3,313 test instances.

\paragraph{Models and Baselines.} As our aim is not to beat previous performances on other benchmarks, but to assess the difficulty of our data set, understand the importance of our gold citation context annotations, and to provide strong baselines for future research, we resort to the most natural model and baseline choices: (1) we report a majority vote baseline, in which we simply predict the majority label. (2)~As domain-specific transformer~\citep{vaswani-attention}-based model, we employ a SciBERT~\citep{beltagy-etal-2019-scibert} encoder, on top of which we place a multi-label classification head consisting of a set of sigmoid classifiers (one for each of the classes) to which we feed the sequence start token. (3)~Last, we replace the encoder with a RoBERTa Model~\citep{liu-roberta}, a more rebustly trained version of BERT~\citep{devlin-etal-2019-bert}. 

\paragraph{Input Preparation.} For the two transformer-based models, we prepare the input as follows: we first highlight the target citation marker by inserting an opening and closing cite tag around the citation, e.g., \texttt{<cite>}Author (Year)\texttt{</cite>}. We next sample $s$ consecutive sentences around the target citation\footnote{The position of the citing sentence within the context varies, to not bias the models towards the mid of the citation context.} as citation context. For instance, for $s=1$, only the citing sentence is sampled and for $s=2$, the citing sentence and the preceding or following sentence is extracted. Alternatively, we employ the annotated gold context. We then follow the standard input procedure for each model, i.e., we apply WordPiece~\citep{johnson2017google} tokenization for SciBERT and byte-level BPE~\citep{radford2019language} for RoBERTa, and add the corresponding special tokens for each model.

\paragraph{Training and Optimization.} SciBERT is only available in base configuration~($12$ layers, $12$ attention heads, $768$ as hidden size, cased vocabulary with size $31,116$). For RoBERTa, we employ the large version ($24$ layers, $16$ attention heads, hidden size $1024$, cased vocabulary with size $50,265$). We conduct standard fine-tuning of the models with a mean over $N$ binary cross-entropy losses:
\begin{equation}
\scriptsize{
L =\frac{1}{N}\sum_{n=1}^{N}-[y_n\cdot \log\hat{y}_n+(1-y_n) \cdot \log(1-\hat{y}_n)]\,,}
\end{equation}
with $N$ as the number of classes, $\hat{y}_n \in \mathbb{R}$ as the sigmoid activated output for class $n$, and $y_n$ as the true label. We optimize using Adam~\citep{kingma2015adam}. To select the best hyperparameters we grid search for the best learning rate $\lambda \in \{1 \cdot 10^{-5}, 2 \cdot 10^{-5}\}$ and number of epochs $e \in [1,9]$ based on the development set performance. The effective batch size is fixed to $32$ and the sigmoid prediction threshold to $0.5$.

\paragraph{Evaluation Measures.} %
We compute two types of accuracies: a \emph{strict} version, in which a prediction is correct iff all predicted labels match exactly the gold annotation; and a \emph{weak} version, in which a prediction is correct if at least one of the predicted labels matches the gold classes. 
The weak measure reflects an upper bound on performance (i.e., whether the model can detect \textit{any} of the correct intents) and allows us to compare our multilabel models with existing single-label models.
Additionally, we break down the performance in different categories according to the gold context size.

\subsection{Results}
\begin{table*}[!t]\centering
\small
\begin{tabularx}{\textwidth}{llXXXXXXXXXXXX}\toprule
\textbf{} &\textbf{} &\multicolumn{2}{c}{\textbf{size = 1}} &\multicolumn{2}{c}{\textbf{size = 2}} &\multicolumn{2}{c}{\textbf{size = 3}} &\multicolumn{2}{c}{\textbf{size = 4}} &\multicolumn{2}{c}{\textbf{all}} \\
& &\multicolumn{2}{c}{support = 2795} &\multicolumn{2}{c}{support = 335} &\multicolumn{2}{c}{support = 112} &\multicolumn{2}{c}{support = 39} &\multicolumn{2}{c}{support = 3313} \\
\cmidrule(lr){3-4}
\cmidrule(lr){5-6}
\cmidrule(lr){7-8}
\cmidrule(lr){9-10}
\cmidrule(lr){11-12}
&input size &weak &strict &weak &strict &weak &strict &weak &strict &weak &strict \\
\midrule
SC Majority & -- &  0.39 & 0.37 & 0.31 & 0.16 & 0.36 & 0.19 & 0.41 & 0.13 & 0.39 &0.34\\
SC Oracle & -- & 1.00 & 0.89 & 1.00 & 0.77 & 1.00 & 0.74 & 1.00 & 0.56 & 1.00 & 0.87 \\
SC ACL-ARC & -- & 0.68 & 0.60 & 0.54 & 0.41 & 0.52 & 0.34 & 0.51 & 0.15  & 0.66 & 0.56\\
\midrule
\multirow{6}{*}{SciBERT} &1 &0.78 &0.69 &0.45 &0.28 &0.47 &0.24 &0.51 &0.18 &0.74 &0.62 \\
&3 &0.74 &0.64 &0.59 &0.39 &0.54 &0.29 &0.62 &0.23 &0.72 &0.60 \\
&5 &0.71 &0.61 &0.50 &0.33 &0.46 &0.27 &0.54 &0.18 &0.68 &0.57 \\
&7 &0.62 &0.54 &0.43 &0.28 &0.48 &0.27 &0.51 &0.15 &0.60 &0.50 \\
&9 &0.56 &0.50 &0.37 &0.25 &0.37 &0.21 &0.56 &0.18 &0.53 &0.46 \\
&gold &\textbf{0.80} &\textbf{0.70} &\textbf{0.68} &\textbf{0.46} &\textbf{0.66} &\textbf{0.39} &\textbf{0.64} &\textbf{0.26} &\textbf{0.78} &\textbf{0.66} \\
\midrule
\multirow{6}{*}{RoBERTa} &1 &0.80 &\textbf{0.69} &0.46 &0.29 &0.46 &0.25 &0.56 &0.18 &0.75 &0.63 \\
&3 &0.78 &0.66 &0.59 &0.41 &0.50 &0.27 &\textbf{0.62} &0.18 &0.75 &0.61 \\
&5 &0.75 &0.63 &0.54 &0.39 &0.54 &0.32 &0.59 &0.21 &0.72 &0.59 \\
&7 &0.73 &0.62 &0.53 &0.37 &0.44 &0.24 &0.56 &0.21 &0.70 &0.58 \\
&9 &0.71 &0.59 &0.54 &0.36 &0.46 &0.26 &0.54 &0.15 &0.68 &0.55 \\
&gold &\textbf{0.81} &\textbf{0.69} &\textbf{0.70} &\textbf{0.50} &\textbf{0.67} &\textbf{0.45} &0.59 &\textbf{0.28} &\textbf{0.79} &\textbf{0.66} \\
\bottomrule
\end{tabularx}
\caption{Results of the multi-label citation context classification with SciBERT and RoBERTa for different input context sizes (input) across all test instances (all) and spread out for different gold context sizes with their respective support in the test set. We report \emph{strict} and \emph{weak} accuracies for gold context sizes with number of supporting instances > $30$. Scores in bold highlight the best performances per model per column. Note that the results for SC ACL-ARC are computed with the reduced label set.}\label{tab:context}
\end{table*}
The results are shown in Table~\ref{tab:context}. We report the results for context sizes with more than $30$ instances in the test set only. %

Across all test instances (all) the best weak and strict accuracies are achieved when feeding the gold context. When employing the citing sentence only, the results drop $4$ percentage points for SciBERT and $3$ (strict) to $4$ (weak) percentage points for RoBERTA. The scores decrease even more, the more sentences are serving as input to the model training with drops up to $25$ percentage points (strict accuracy for SciBERT trained on 9-sentence contexts). When categorizing the prediction instances according to their gold context size, more interesting patterns emerge: for contexts consisting of the citing sentence only, the delta between the model trained on gold context and the model trained on the citing sentence is naturally smaller, but surprisingly, they seem to be still existent ($2$ percentage points for weak, $1$ percentage point for strict accuracy with SciBERT; $1$ percentage point for weak accurcy with RoBERTa). We hypothesize that this indicates that the models trained on the gold contexts are able to learn more about the intent classes than the single-sentence models. This, again, highlights the importance of considering precisely-sized contexts in training, even if the majority of the prediction instances are single-sentence instances only. For gold context sizes $3$ and $4$, the gap between the models trained on gold contexts and the ones trained on the citing sentence increases even more, but we can also see that models trained on $3$ and $4$ instances are better able to capture true intents than the $1$-sentence models.
\section{Citation Analysis as Question Answering}
\label{qa}
Consider the following application scenario: a user might want to know why a paper cites another paper or for which reasons a paper is referenced. In both cases, they expect a paper-level evaluation (i.e., the model should retrieve a set of intents), and, at the same time, they would want to see evidence supporting the results (the model should retrieve a gold context for each intent). We propose to fulfill these two desiderata by resorting to a question answering~(QA) formulation as another instance of our Multi-sentence Multi-intent framework. The advantages of this approach are (a) its flexibility, as, theoretically, users can any questions, and (b) its compatibility with general attempts on scientific QA~\citep[e.g.,][]{dasigi-etal-2021-dataset}. Further, this QA reframing allows CCA to be solved using QA models, treating CCA as a challenging case of natural language understanding in the scientific domain.

\subsection{Task Formulation and Methodology}
We adapt the Qasper~\citep[e.g.,][]{dasigi-etal-2021-dataset} task formulation and model for our purposes. Qasper is a document-grounded QA model that has been pre-trained to answer a variety of questions on scientific texts, e.g., extractive QA or yes/no questions, making it ideal as an initial model to adapt for CCA.

\paragraph{Task.} The task is formulated as follows: given a paper pair ($A$ and $B$, as before) and a question related to the intent of $A$ citing $B$, output the answer based on the full text of $A$ as well as an evidence for the given answer, i.e., a single gold context $E= [s_{e0},...,s_{eO}]$ as a (dis)continuous sequence of sentences $s_i$ from $A$. Concretely, we ask binary \emph{yes/no}-questions for each of the intents, e.g., for the \emph{background} intent, we ask \emph{``Does the paper cite ... for background information?''} (dots are replaced with the respective citation marker).

\paragraph{Model.} We use the Qasper evidence scaffold model: this model is a multi-task sequence-to-sequence Longformer-Encoder-Decoder~\citep[LED;][]{Beltagy2020Longformer} model. LED is a variant of the Transformer~\citep{vaswani-attention} encoder-decoder model which supports processing long inputs, e.g., scientific papers, due to a sparse attention mechanism which scales linearly with the input sequence length. The original LED parameters are initialized from BART~\citep{lewis-etal-2020-bart}.
As model input, the question and paper context are concatenated to a single string. We prepend each sentence in the context with a \texttt{</s>} token and globally attend over all question tokens and \texttt{</s>} tokens. 
The model objective consists of two parts: a generative answering component, and an evidence extraction component. 
(1) The generative answering component is a classifier over the model's vocabulary trained with a cross-entropy loss to generate \emph{``Yes''} or \emph{``No''}.\footnote{Using the generative component as in the original work allows us to (a) keep the flexibility of generating other answers and (b) reuse Qasper's head weights if desired.} (2) The evidence extraction component is a classifier over the \texttt{</s>} tokens, trained with a binary cross-entropy loss for evidence/non-evidence sentences. As the class of positive, i.e., evidentiary, sentences is underrepresented in the data set, we follow the original work and scale the loss proportional to the ratio of positive/negative gold sentences per batch. The total loss corresponds to the sum of the two task losses.

\subsection{Experimental Setup}
We describe the experimental setup of our QA experiments.

\paragraph{Data.} We use the same standard split as in our classification experiments. However, we preprocess the data according the model's required input format, i.e., we create for each paper-pair seven questions (one for each of our seven intents). For all positive intents, we create a \emph{``Yes''}-answer and provide the first gold context as evidence. For the negative intents, we create a \emph{``No''}-answer without evidence. This way, we end up with 4,074 training, 1,764 development, and 2,499 test questions.

\paragraph{Model Configuration and Optimization.} We employ two LED models: first, we start from the original LED base model (12 attention heads and 6 layers in encoder and decoder, respectively; 768 as hidden size; 50,265 as vocabulary size, maximum input length 16,384 tokens). Secondly, to estimate the complementarity of the knowledge needed for CCA QA with more general scientific QA as in Qasper, we start from the LED base model trained on the Qasper data set shared by the authors of the paper (encoder only). For all models, we use the code from \citet{dasigi-etal-2021-dataset}. We train all models for maximum $5$ epochs with early stopping based on the validation set performance (span-level Answer-F1 as described below, patience of 2 epochs) and grid search over the following hyperparameters: batch size $b \in \{2, 4, 8, 16\}$ and initial learning rate $\lambda \in \{3 \cdot 10^{-5}, 5 \cdot 10^{-5}\}$. We optimize all models with Adam~\citep{kingma2015adam}.

\paragraph{Evaluation Measures and Baselines.} %
For evaluating answer performance, we follow \citet{dasigi-etal-2021-dataset} and report a binary F1 measure micro averaged across the papers. Similarly, to evaluate the evidence extraction performance, we compute an evidence F1 where we compare the gold context for an answer with the per-sentence predictions of the model. %
Additionally, as for all \emph{``No''}-answers, i.e., for all citation intents which do not apply, also no evidence is given, we compute an F1 only considering the positive intents, dubbed \emph{Evidence F1 w/o No}. Similarly, we compute an additional evidence score, which we condition on the correct model answer predictions only (\emph{Evidence F1 correct}).
To estimate the difficulty of the task, we report a two simple majority vote baselines. In the first variant, dubbed \emph{Majority SL}, we only predict \emph{``yes''} for the majority class, \emph{Background}. In the second variant, to which we refer to as \emph{Majority ML}, we compute for each question type each of which relates to a single intent, the majority answer (\emph{``yes''} or \emph{``no''}). %

\subsection{Results} 

\begin{table}[!t]\centering
\small
\begin{tabularx}{\linewidth}{lXXXXX}\toprule
 & \textbf{A F1} & \textbf{E F1} & \textbf{E F1 w/o No} & \textbf{E F1 correct}\\\midrule
Majority SL & 0.61 & 0.48 & 0.00 & 0.47 \\
Majority ML & 0.72 & 0.48 & 0.00 & 0.38 \\
\midrule
LED Qasper & 0.75 & 0.48 & 0.00 & 0.35 \\
LED Base & 0.78 &  0.08 & 0.04 & 0.07\\
\bottomrule
\end{tabularx}
\caption{Citation Context Analysis QA results. We report Answer-F1 (A F1), Evidence-F1 (E F1), as well as its variants Evidence-F1 w/o No (E F1 w/o No) and Evidence-F1 for correct answers only (E F1 correct) .}\label{tab:qa}
\end{table}
The results are shown in Table~\ref{tab:qa}. Both LED models surpass the baselines in terms of Answer F1. Surprisingly, the knowledge from the Qasper LED encoder does not seem to lead to strong performance gains. Relating to the Evidence F1 scores, interesting patterns emerge: overall the scores from LED Qasper are comparable to the ones reported in the original work for general scientific QA. However, when negative answers are not considered in the LED Qasper model's performance (excluding contexts which no answer should be given), the resulting score shows that the model does not properly extract context sentences. The LED Base model, in contrast, learns to extract context, though the scores are very low. We attribute this observation to the amount of context given.
Overall, the results indicate the difficulty of the task under this natural formulation and we propose that our initial experiments open up an interesting and challenging research avenue for QA in the context of CCA.

\section{Conclusion}
In this work, we have presented a new Multi-Sentence Multi-Intent CCA framework. In a qualitative analysis of citation contexts, we demonstrated the importance of considering multiple sentences and multiple labels for citation intent classification. In lack of a data set which allows to operate under our new framework, we presented \corpus{}, an annotated corpus consisting of 12,653 citations across 1,193 fully-annotated NLP publications. Next, we employed \corpus{} in traditional classification experiments, which showed the importance of using our annotated gold contexts as model inputs. %
Finally, we proposed to cast citation context analysis as question answering task, which allows for more flexibility and end-to-end modeling of the problem. This way, the task integrates with other reading comprehension tasks on scientific publications. We hope that our work draws more attention to research on previously overlooked phenomena in citation context analysis to support more accurate research evaluation studies.

\section*{Acknowledgements}
We would like to thank Sam Skjonsberg and Mark Neumann for their help on the annotation interface.  We would also like to thank Noah Smith for helping us connect with collaborators on this project.

\bibliography{custom}
\bibliographystyle{acl_natbib}

\setlength{\tabcolsep}{2.4pt}
\begin{table*}[ht!]\centering
\scriptsize
\begin{tabularx}{\textwidth}{llXXXXXXXXXXXXXXXXXXXXXXXX}\toprule
\textbf{} &\textbf{} &\multicolumn{2}{c}{\textbf{size = 1}} &\multicolumn{2}{c}{\textbf{size = 2}} &\multicolumn{2}{c}{\textbf{size = 3}} &\multicolumn{2}{c}{\textbf{size = 4}} &\multicolumn{2}{c}{\textbf{size = 5}} &\multicolumn{2}{c}{\textbf{size = 6}} &\multicolumn{2}{c}{\textbf{size = 7}} &\multicolumn{2}{c}{\textbf{size = 8}} &\multicolumn{2}{c}{\textbf{size = 10}} &\multicolumn{2}{c}{\textbf{size = 14}} &\multicolumn{2}{c}{\textbf{all}} \\
\textbf{} &\textbf{} &\multicolumn{2}{c}{supp. = 2795} &\multicolumn{2}{c}{supp. = 335} &\multicolumn{2}{c}{supp. = 112} &\multicolumn{2}{c}{supp. = 39} &\multicolumn{2}{c}{supp. = 17} &\multicolumn{2}{c}{supp. = 7} &\multicolumn{2}{c}{supp. = 3} &\multicolumn{2}{c}{supp. = 2} &\multicolumn{2}{c}{supp. = 2} &\multicolumn{2}{c}{supp. = 1} &\multicolumn{2}{c}{supp. = 3313} \\ 
\textbf{} &\textbf{Input} &weak &strict &weak &strict &weak &strict &weak &strict &weak &strict &weak &strict &weak &strict &weak &strict &weak &strict &weak &strict &weak &strict \\
\cmidrule(lr){3-4}
\cmidrule(lr){5-6}
\cmidrule(lr){7-8}
\cmidrule(lr){9-10}
\cmidrule(lr){11-12}
\cmidrule(lr){13-14}
\cmidrule(lr){15-16}
\cmidrule(lr){17-18}
\cmidrule(lr){19-20}
\cmidrule(lr){21-22}
\cmidrule(lr){23-24}
S &1 &0.78 &0.69 &0.45 &0.28 &0.47 &0.24 &0.51 &0.18 &0.59 &0.29 &\textbf{0.86} &0.57 &\textbf{1.00} &\textbf{0.67} &\textbf{1.00} &0.50 &\textbf{0.50} &0.00 &0.00 &0.00 &0.74 &0.62 \\
&3 &0.74 &0.64 &0.59 &0.39 &0.54 &0.29 &0.62 &0.23 &\textbf{0.65} &\textbf{0.35} &\textbf{0.86} &0.57 &\textbf{1.00} &\textbf{0.67} &\textbf{1.00} &0.50 &\textbf{0.50} &0.00 &0.00 &0.00 &0.72 &0.60 \\
&5 &0.71 &0.61 &0.50 &0.33 &0.46 &0.27 &0.54 &0.18 &0.53 &0.24 &\textbf{0.86} &0.57 &\textbf{1.00} &\textbf{0.67} &\textbf{1.00} &0.50 &\textbf{0.50} &0.00 &0.00 &0.00 &0.68 &0.57 \\
&7 &0.62 &0.54 &0.43 &0.28 &0.48 &0.27 &0.51 &0.15 &0.53 &0.24 &0.57 &0.43 &\textbf{1.00} &\textbf{0.67} &\textbf{1.00} &0.50 &\textbf{0.50} &0.00 &0.00 &0.00 &0.60 &0.50 \\
&9 &0.56 &0.50 &0.37 &0.25 &0.37 &0.21 &0.56 &0.18 &0.53 &0.29 &0.71 &0.57 &\textbf{1.00} &\textbf{0.67} &0.50 &0.50 &\textbf{0.50} &0.00 &0.00 &0.00 &0.53 &0.46 \\
&gold &\textbf{0.80} &\textbf{0.70} &\textbf{0.68} &\textbf{0.46} &\textbf{0.66} &\textbf{0.39} &\textbf{0.64} &\textbf{0.26} &\textbf{0.65} &0.24 &\textbf{0.86} &\textbf{0.86} &\textbf{1.00} &\textbf{0.67} &\textbf{1.00} &\textbf{1.00} &\textbf{0.50} &0.00 &\textbf{1.00} &0.00 &\textbf{0.78} &\textbf{0.66} \\
\midrule
R &1 &0.80 &\textbf{0.69} &0.46 &0.29 &0.46 &0.25 &0.56 &0.18 &0.65 &0.35 &\textbf{0.86} &0.57 &\textbf{1.00} &\textbf{0.67} &0.50 &0.50 &0.50 &0.00 &0.00 &0.00 &0.75 &0.63 \\
&3 &0.78 &0.66 &0.59 &0.41 &0.50 &0.27 &0.62 &0.18 &\textbf{0.76} &\textbf{0.41} &\textbf{0.86} &0.71 &\textbf{1.00} &\textbf{0.67} &\textbf{1.00} &\textbf{1.00} &0.50 &0.00 &0.00 &0.00 &0.75 &0.61 \\
&5 &0.75 &0.63 &0.54 &0.39 &0.54 &0.32 &\textbf{0.59} &0.21 &0.65 &\textbf{0.41} &0.71 &0.57 &\textbf{1.00} &\textbf{0.67} &\textbf{1.00} &0.50 &0.50 &0.00 &0.00 &0.00 &0.72 &0.59 \\
&7 &0.73 &0.62 &0.53 &0.37 &0.44 &0.24 &0.56 &0.21 &0.53 &0.35 &0.71 &0.57 &\textbf{1.00} &\textbf{0.67} &0.50 &0.50 &0.50 &0.00 &0.00 &0.00 &0.70 &0.58 \\
&9 &0.71 &0.59 &0.54 &0.36 &0.46 &0.26 &0.54 &0.15 &0.65 &0.35 &\textbf{0.86} &\textbf{0.86} &\textbf{1.00} &\textbf{0.67} &0.50 &0.50 &0.50 &0.00 &0.00 &0.00 &0.68 &0.55 \\
&gold &\textbf{0.81} &\textbf{0.69} &\textbf{0.70} &\textbf{0.50} &\textbf{0.67} &\textbf{0.45} &\textbf{0.59} &\textbf{0.28} &0.71 &0.35 &\textbf{0.86} &\textbf{0.86} &0.67 &0.33 &\textbf{1.00} &\textbf{1.00} &\textbf{1.00} &\textbf{0.50} &0.00 &0.00 &\textbf{0.79} &\textbf{0.66} \\
\bottomrule
\end{tabularx}
\caption{Full results of the multi-label citation context classification with SciBERT (S) and RoBERTa (R) for different input context sizes (input) across all test instances (all) and spread out for different gold context sizes with their respective support in the test set. We report \emph{strict} and \emph{weak} accuracies. Scores in bold highlight the best performances per column.}\label{tab:app_context}
\end{table*}

\FloatBarrier       %
\appendix

\section{Full Results for Section 5}
\label{sec:appendix}

We provide the full experimental results for the classification experiments described in Section 5 in Table~\ref{tab:app_context}.

\end{document}